\documentclass[wcp]{jmlr}

\usepackage{longtable}

\usepackage{booktabs}

\usepackage{bm}

\usepackage{times}
\usepackage{graphicx}
\usepackage{latexsym}

\usepackage{times}
\usepackage{multirow}

\usepackage{graphicx}

\usepackage{amssymb}
\usepackage{algorithm}
\usepackage{algorithmic}
\usepackage{array}

\usepackage{fancyvrb}
\usepackage{amsfonts}
\usepackage{times}
\usepackage{helvet}
\usepackage{courier}
\usepackage{graphics}
\usepackage{colordvi,multirow,amsmath,amssymb}

\DeclareMathOperator*{\argmax}{arg\,max}

\jmlrvolume{60}
\jmlryear{2016}
\jmlrworkshop{ACML 2016}

\newcommand{\nop}[1]{}

\newcommand{\bx}{\textbf{x}}
\newcommand{\bs}{\textbf{s}}

\newcommand{\mT}[0]{\mathcal{T}}
\newcommand{\mX}[0]{\mathcal{X}}
\newcommand{\mO}[0]{\mathcal{O}}
\newcommand{\mS}[0]{\mathcal{S}}
\newcommand{\mP}[0]{\mathcal{P}}

\newtheorem{defn}{Definition}

\title{Extracting Actionability from Machine Learning Models by Sub-optimal Deterministic Planning}

\author{\Name{Qiang Lyu},
 \Name{Yixin Chen}, 
\Name{Zhaorong  Li}, 
\Name{Zhicheng Cui},  
\Name{Ling  Chen},
\Name{Xing   Zhang}, 
\Name{Haihua   Shen}
}

\begin{document}
\maketitle

\begin{abstract}
A main focus of machine learning  research has been improving the generalization accuracy and efficiency of prediction models. Many models such as SVM, random forest, and deep neural nets have been proposed and achieved great success. However, what emerges as missing in many applications is actionability, i.e., the ability to turn prediction results into actions. For example, in applications such as customer relationship management, clinical prediction, and advertisement, the users need not only accurate prediction, but also actionable instructions which can transfer an input to a desirable goal (e.g., higher profit repays, lower  morbidity rates, higher ads hit rates). Existing effort in deriving such actionable knowledge is few and limited to simple action models which restricted to only change one attribute for each action. The dilemma is that in many real applications those action models are often more complex and harder to extract an optimal solution.

In this paper, we propose a novel approach that achieves actionability by combining learning with planning, two core areas of AI. In particular, we propose a framework to extract actionable knowledge from random forest, one of the most widely used and best off-the-shelf classifiers. We formulate the actionability problem to a sub-optimal action planning (SOAP) problem, which is to find a plan to alter certain features of a given input so that the random forest would yield a desirable output, while minimizing the total costs of actions. Technically, the SOAP problem is formulated in the SAS+ planning formalism, and solved using a Max-SAT based approach. Our experimental results demonstrate the effectiveness and efficiency of the proposed approach on a personal credit dataset and other benchmarks. Our work represents a new application of automated planning on an emerging and challenging machine learning paradigm.

\end{abstract}

\begin{keywords}
actionable knowledge extraction, machine learning, planning, random forest, weighted partial Max-SAT
\end{keywords}




\section{Introduction}

Research on machine learning has achieved great success on enhancing the models' accuracy and efficiency. 
Successful models such as support vector machines (SVMs), random forests, and deep neural nets have been applied to vast industrial applications~\cite{mitchell1999machine}.
However, in many applications, users may need not only a prediction model, but also suggestions on courses of actions to achieve desirable goals.
For practitioners, a complex model such as a random forest is often not very useful even if its accuracy is high because of its lack of actionability.
Given a learning model, extraction of actionable knowledge entails finding a set of actions to change the input features of a given instance so that it achieves a desired output from the learning model. We elaborate this problem using one example.




\noindent{\bf Example 1}. In a credit card company, a key task is to decide on promotion strategies to maximize the long-term profit.
The customer relationship management (CRM) department collects data about customers, such as customer education, age, card type, the channel of initiating the card, the number and effect of different kinds of promotions, the number and time of phone contacts, etc.

For data scientists, they need to build models to predict the profit brought by customers.
In a real case, a company builds a random forest involving 35 customer features. The model predicts the profit (with  probability) for each customer.
In addition, a more important task is to extract actionable knowledge to revert ``negative profit'' customers and retain  ``positive profit'' customers.
In general, it is much cheaper to maintain existing ``positive profit''customers than to revert  ``negative profit'' ones. It is especially valuable to retain high profit, large, enterprise-level customers.

There are certain actions that the  company can take, such as making phone contacts and sending promotional coupons.
Each action can change the value of one or multiple attributes of a customer.
Obviously, such actions incur costs for the company.
For instance, there are 7 different kinds of promotions and each promotion associates with two features, the number and the accumulation effect of sending this kind of promotion.
When performing an action of ``sending promotion\_amt\_N'', it will change features ``nbr\_promotion\_amt\_N'' and ``s\_amt\_N'', the number and the accumulation effect of sending the sales promotion, respectively.
For a customer with ``negative profit'', the goal is to extract a sequence of actions that change the customer profile so that the model gives a ``positive profit" prediction while minimizing the total action costs.
For a customer with ``positive profit'', the goal is to find actions so that the customer has a  ``positive profit'' prediction with a higher prediction probability. $\blacksquare$


Research on extracting actionability from machine learning models is still limited. There are a few existing works.
Statisticians have adopted stochastic models to find specific rules of the response behavior of customer~\cite{desarbo1994crisp,levin1996segmentation}.
There have also been efforts on the development of ranking mechanisms with business interests~\cite{Hilderman00,Cao07} and pruning and summarizing learnt rules by considering similarity~\cite{Liu96,Liu99,cao2007domain,cao2010flexible}.
However, such approaches are not suitable for the problems studied in this paper due to two major drawbacks.
First, they can not provide customized actionable knowledge for each individual since the rules or rankings are derived from the entire population of training data.
Second, they did not consider the action costs while building the rules or rankings.
For example, a low income housewife may be more sensitive to sales promotion driven by consumption target, while a social housewife may be more interested in promotions related to social networks.
Thus, these rule-based and ranking algorithms cannot tackle these problems very well since they are not personalized for each customer.

Another related work is extracting actionable knowledge from decision tree and additive tree models by bounded tree search and integer linear programming~\cite{Yang03,yang2007extracting,cui2015}.
Yang's work focuses on finding optimal strategies by using a greedy strategy to search on one or multiple  decision trees~\cite{Yang03,yang2007extracting}.
Cui et al. use an integer linear programming (ILP) method to find actions changing sample membership on an ensemble of trees~\cite{cui2015}.
A limitation of these works is that the actions are assumed to change only one attribute each time.
As we  discussed above, actions like ``sending promotion\_amt\_N'' may change multiple features, such as  ``nbr\_promotion\_amt\_N'' and ``s\_amt\_N''. Moreover, Yang's greedy method is fast but cannot give optimal solution~\cite{Yang03}, and Cui's optimization method is optimal but very slow~\cite{cui2015}.

In order to address these challenges, we propose a novel approach to extract actionable knowledge from random forests, one of the most popular learning models. Our approach leverages planning, one of the core and extensively researched areas of AI.
We first rigorously formulate the knowledge extracting problem to a sub-optimal actionable planning (SOAP) problem which is defined as finding a sequence of actions transferring a given input to a desirable goal while minimizing the total action costs. Then, our approach consists of two phases.
In the offline preprocessing phase, we use an anytime state-space search on an action graph to find a preferred goal for each instance in the training dataset and store the results in a database.
In the online phase, for any given input, we translate the SOAP problem into a SAS+ planning problem.
The SAS+ planning problem is solved by an efficient MaxSAT-based approach  capable of optimizing plan metrics. 

We perform empirical studies to evaluate our approach. We use a real-world credit card company dataset obtained through an industrial research collaboration. We also evaluate some other standard benchmark datasets. We compare the quality and efficiency of our method to several other state-of-the-art methods.
The experimental results show that our method achieves a near-optimal quality and real-time online search as compared to other existing methods.


\section{Preliminaries}

\subsection{Random forest}
\label{sec:randomforest}

Random forest  is a popular model for classification, one of the main tasks of learning.
The reasons why we choose Random forest are: 1) In addition to superior classification/regression performance, Random forest enjoys many appealing properties many other models lack~\cite{friedman2001elements}, including the support for multi-class classification and natural handling of missing values and data of mixed types.
2) Often referred to as one of the best off-the-shelf classifier~\cite{friedman2001elements}, Random forest has been widely deployed in many industrial products such as Kinect~\cite{shotton2013real} and face detection in camera~\cite{viola2004robust}, and is the popular method for some competitions such as web search ranking~\cite{mohan2011web}.





Consider a dataset $\{X,Y\}$, where $X=\{{\bx}^1, \cdots, {\bx}^N\}$ is the set of training samples and $Y=\{y^1, \cdots, y^N\}$ is the set of classification labels.
Each vector ${\bx}^i=(x^i_1, \cdots, x^i_M)$ consists of $M$ attributes, where each attribute $x_j$ can be either categorical or numerical and has a finite or infinite domain $Dom(x_j)$.
Note that we use ${\bx}=(x_1, \cdots, x_M)$ to represent ${\bx}^i =(x^i_1, \cdots, x^i_M)$ when there is no confusion.
All labels $y^i$ have the same finite categorical domain $Dom(Y)$.

A random forest contains $D$ decision trees where each decision tree $d$ takes an input ${\bx}$ and outputs a label $y\in Dom(Y)$, denoted as $o_d({\bx})=y$.
For any label $c \in Dom(Y)$, the probability of output $c$ is
\begin{equation} \label{eq:prob}
p\Big(y=c \mid \bx\Big) = \frac{\sum_{d=1}^{D} w_d I(o_d({\bx}) = c)}{\sum_{d=1}^{D} w_d},
\end{equation}
where $w_d \in \mathcal{R}$ are weights of decision trees, $I(o_d({\bx})= c)$ is an indicator function which evaluates to 1 if $o_d({\bx})= c$ and 0 otherwise.
The overall output predicted label is 
\begin{equation} \label{eq:H}
H(\bx) = \argmax_{c \in Dom(Y)}~~ p(y=c | \bx).
\end{equation}








A random forest is generated as follows~\cite{breiman2001random}.
For $d=1, \cdots, D$,
\begin{enumerate}
\item Sample $n_k$ ($0<n_k<N$)  instances from the dataset with replacement.


\item Train an un-pruned decision tree on the $n_k$ sampled instances. At each node, choose the split point from a number of randomly selected features rather than all features.
\end{enumerate}


\subsection{SAS+ formalism}
\label{subsec:sas}

In classical planning, there are two popular formalisms, STRIPS and PDDL~\cite{PDDL21}. 
In recent years, another indirect formalism, SAS+, has attracted increasing uses due to its many favorable features, such as compact encoding with multi-valued variables,  natural support for invariants, associated domain transition graphs (DTGs) and causal graphs (CGs) which capture vital structural information~\cite{backstrom1995complexity,jonsson&Bm98,helmert2006fast}.

In SAS+ formalism, a planning problem is defined over a set of multi-valued \emph{state variables} $\mX=\{x_1,\cdots, x_{|\mX|}\}$.
Each variable $x \in \mX$ has a finite domain $Dom(x)$.
A state $s$ is a full assignment of all the variables.
If a variable $x$ is assigned to $f \in Dom(x)$ at a state $s$, we denote it as $s(x)=f$.
We use $\mS$ to represent the set of all states.

\begin{defn} \label{defn:sas-transition} \textbf{(Transition)} Given a multi-valued state variable $x\in \mX$ with a domain $Dom(x)$, a transition is defined as a tuple $\mT=(x, f, g)$, where $f,g\in Dom(x)$, written as $\delta^{x}_{f\rightarrow g}$. A transition $\delta^{x}_{f\rightarrow g}$ is applicable to a state $s$ if and only if $s(x)=f$.
We use $\oplus$ to represent applying a transition to a state.
Let $s'=s\oplus \delta^{x}_{f\rightarrow g}$ be the state after applying the transition to $s$, we have $s'(x)=g$.
We also simplify the notation $\delta^{x}_{f\rightarrow g}$ as $\delta^{x}$ or $\delta$ when there is no confusion.
\end{defn}

A transition $\delta^{x}_{f\rightarrow g}$ is a \textbf{regular transition} if $f\neq g$ or a \textbf{prevailing transition} if $f = g$. In addition,
 $\delta^{x}_{*\rightarrow g}$ denotes a \textbf{mechanical transition}, which can be applied to any state $s$ and changes the value of $x$ to $g$.

\nop{According to the values of $f$ and $g$, SAS+ transitions can be classified into three categories.
\begin{itemize}
\item \textbf{Regular transition}. A transition $\delta^{x}_{f\rightarrow g}$ is a regular transition if and only if $f\neq g$.  
\item \textbf{Prevailing transition}.  A transition $\delta^{x}_{f\rightarrow g}$ is a prevailing transition if and only if $f = g$.  
\item \textbf{Mechanical transition}. Transitions of the form $\delta^{x}_{*\rightarrow g}$ are called mechanical. A mechanical transition $\delta^{x}_{*\rightarrow g}$ can be applied to an arbitrary state $s$.
\end{itemize}
}

For a  variable $x$, we denote the set of all transitions that affect $x$ as $\mT(x)$, i.e., $\mT(x)=\{\delta^{x}_{f\rightarrow g}\}\cup \{\delta^{x}_{*\rightarrow g}\}$
for all $f,g\in Dom(x)$.
We also denote the set of all transitions as $\mT$, i.e., $\mT=\bigcup_{x \in \mX}{\mT(x)}$.

\begin{defn} \label{defn:sas-mutex} \textbf{(Transition mutex)} For two different transitions $\delta^{x}_{f\rightarrow g}$ and $\delta^{x}_{f'\rightarrow g'}$, if
at least one of them is a mechanical transition and $g=g'$, they are compatible; otherwise, they are
mutually exclusive (mutex).
\end{defn}

\begin{defn} \label{defn:sas-action} \textbf{(Action)} An action $a$ is a set of transitions $\{\delta_1, \cdots, \delta_{|a|}\}$, where there do not exist two transitions $\delta_i, \delta_j \in a$ that are mutually exclusive. An action $a$ is \textbf{applicable} to a state $s$ if and only if all transitions in $a$ are applicable to $s$. Each action has a \textbf{cost} $\pi(a) > 0$.
\end{defn}

\begin{defn} \label{def:sas-problem} \textbf{(SAS+ planning)}
A SAS+ planning problem is a tuple $\Pi_{sas}=(\mathcal{X}, \mathcal{O}, s_I, S_G)$ defined as follows
\begin{itemize}
\item $\mathcal{X}=\{x_1,\cdots, x_{|\mX|}\}$ is a set of state variables.
\item $\mathcal{O}$ is a set of actions.
\item $s_I \in \mS$ is the initial state.
\item $S_G$ is a set of goal conditions, where each goal condition $s_G \in S_G$ is a partial assignment of some state variables. A state $s$ is a goal state if there exists $s_G \in S_G$ such that $s$ agrees with every variable assignment in $s_G$.
\end{itemize}
\end{defn}

Note that we made a slight generalization of original SAS+ planning, in which $S_G$ includes only one goal condition. For a state $s$ with an applicable action $a$, we use $s'=s \oplus a$ to denote the resulting state after applying all the transitions in $a$ to $s$ (in an arbitrary order since they are mutex free).



\begin{defn} \label{defn:sas-action-mutex} \textbf{(Action mutex)} Two different actions $a_1$ and $ a_2$ are mutually exclusive if and only if at least one of the following conditions is satisfied:
\begin{itemize}
\item There exists a non-prevailing transition $\delta$ such that $\delta \in a_1$ and $\delta \in a_2$.
\item There exist two transitions $\delta_1 \in a_1$ and $\delta_2 \in a_2$ such that $\delta_1$ and $\delta_2$ are mutually exclusive.
\end{itemize}
\end{defn}

A set of actions $P$ is applicable to $s$ if each action $a \in P$ is applicable to $s$ and no two actions  in $P$ are mutex.
We denote the resulting state after applying a set of actions $P$ to $s$ as $s'=s \oplus P$.

\begin{defn} \label{defn:sas-plan} \textbf{(Solution plan)} For a SAS+ problem $\Pi_{sas}=$\\$(\mathcal{X}, \mathcal{O}, s_I, S_G)$, a solution plan is a sequence  $\mP=(P_1, \cdots, P_L)$, where each $P_t$, $t = 1, \cdots, L$ is a set of actions, and there exists $s_G \in S_G$,
$s_G \subseteq s_I \oplus P_1 \oplus P_2 \oplus \cdots P_L$.
\end{defn}

Note that in a solution plan, multiple non-mutex actions can be applied at the same time step.
$s \oplus P_t$ means applying all actions in $P_t$ in any order to state $s$.
In this work, we want to find a solution plan that minimizes a quality metric, the \textbf{total action cost} $\sum_{P_t \in \mP} \sum_{a \in P_t} \pi(a)$.

\nop{
 There are two typical quality metrics for a solution plan: \\
  1) total action cost $\sum_{a \in \mP} \pi(a)$, \\
  2) makespan ($L$, the number of time steps in $\mP$).
}

\section{Sub-Optimal Actionable Plan (SOAP) Problem}
\label{sec:OAP}

We first give an intuitive description of the SOAP problem.
Given a random forest and an input $\bx$, the SOAP problem is to find a sequence of actions that, when applied to $\bx$, changes it to a new instance which has a desirable output label from the random forest.
Since each action incurs a cost, it also needs to minimize the total action costs.
In general, the actions and their costs are determined by domain experts. For example, analysts in a credit card company can decide  which actions they can perform and how much each action costs.
 
 There are two kinds of features, \emph{soft attributes} which can be changed with reasonable costs and \emph{hard attributes} which cannot be changed with a reasonable cost, such as gender~\cite{Yang03}. We only consider actions that change soft attributes.


\begin{defn} \label{def:OAP} \textbf{(SOAP problem)}  A SOAP problem is a tuple $\Pi_{soap}=(H, {\bx}^I, c, O)$, where $H$ is a random forest, $\bx^I$ is a given input,  $c \in Dom(Y)$ is a class label, and $O$ is a set of actions. The goal is to find a sequence of actions $A=(a_1, \cdots, a_n), a_i \in O$,  to solve:
\begin{eqnarray}
\min_{A \subseteq O} && F(A) = \sum_{a_i \in A } \pi(a_i), \label{eq:profit}\\
\textrm{subject to:} &&  p(y=c|\tilde{\bx}) \ge z, \label{eq:constraint}
\end{eqnarray}
where $\pi(a) > 0$ is the cost of action $a$, $0 < z \le 1$ is a constant, $p(y=c|\tilde{\bx})$ is the output of $H$ as defined in (\ref{eq:prob}), and
$\tilde{\bx}  = \bx^I \oplus a_1 \oplus a_2 \oplus \cdots \oplus a_n$
is the new instance after applying the actions in $A$ to $\bx^I$.
\end{defn}


\begin{figure}[t]
    \centering
    \includegraphics[scale=0.35]{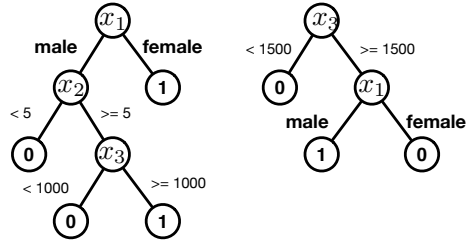}
    \caption{An illustration of a random forest}
    \label{fig:forest}
\end{figure}

\noindent\textbf{Example 2.} A random forest $H$ with two trees and three features is shown in Figure~\ref{fig:forest}.
$x_1$ is a hard attribute, $x_2$ and $x_3$ are soft attributes.
Given $H$ and an input $\bx=(male, 2, 500)$, the output from $H$ is 0. The goal is to change $\bx$ to a new instance that has an output  of 1 from $H$.
For example, two actions changing $x_2$ from 2 to 5 and $x_3$ from 500 to 1500 is a plan and the new instance is $(male, 5, 1500)$. $\blacksquare$


\nop{
In this paper, we first propose an optimal algorithm based on best-first state space search which can guarantee to find the optimal solution without considering the search time limit.
However, the optimal algorithm is very time consuming and sometimes can not stop by proving a found solution is optimal.
Thus, we propose an enhanced greedy algorithm which can solve the problem efficiently and find a near optimal solution.
}

\section{A Planning Approach to SOAP}
\label{sec:two-step-framewrok}


The SOAP problem is proven to be an NP-hard problem, even when an action can change only one feature~\cite{cui2015}. Therefore, we cannot expect any efficient algorithm for optimally solving it.
We propose a planning-based approach to solve the SOAP problem. 
Our approach consists of  an offline preprocessing phase that only needs to be run once for a given random forest, and an online phase that is used to solve each SOAP problem instance.


\subsection{Action graph and preferred goals}
\label{subsec:offline}

Since there are typically prohibitively high number of possible instances in the feature space, it is too expensive and unnecessary to explore the entire space. We reason that the training dataset for building the random forest gives a representative distribution of the instances. Therefore, in the offline preprocessing, we form an action graph and identify a preferred goal state for each training sample.

\begin{defn} \textbf{(Feature partitions)}
Given a random forest $H$, we split the domain of each feature $x_i$ ($i=1, \cdots, M$) into a number of partitions according to the following rules.
 \begin{enumerate}
\item  $x_i$ is split into $n$ partitions if $x_i$ is categorical and has $n$ categories.
\item $x_i$ is split into $n+1$ partitions if $x_i $ is numerical and has $n$ \textbf{branching nodes} in all the decision trees in $H$.
Suppose the branching nodes are $(b_1,\cdots,b_{n})$, the partitions are\\ $\{(-\infty,b_1), [b_1,b_2),\cdots,[b_n,+\infty)\}$.
\end{enumerate}
\end{defn}

In Example 2, $x_1$ is splited into $\{male,female\}$, $x_2$ and $x_3$ are splited into $\{(-\infty, 5),[5, +\infty)\}$ and $\{(-\infty, 1000),[1000, 1500), [1500, +\infty)\}$, respectively.

\begin{defn} \label{defn:state-transformation} \textbf{(State transformation)}
For a given instance $\bx = (x_1, \cdots, x_M)$, let $n_i$ be the number of partitions and $p_i$ the partition index for feature $x_i$, we transform it to a SAS+ state $s(\bx)=(z_1, \cdots, z_M)$, where $|Dom(z_i)|=n_i$ and $s(z_i)=p_i, i=1, \cdots, M$.
\end{defn}

For simplicity, we use $s$ to represent $s(\bx)$ when there is no confusion.
Note that if two instances $\bx_1$ and $\bx_2$ transform to the same state $s$, then they have the same output from the random forest since they fall within the same partition for every feature. In that case, we can use $p(y=c|s)$ in place of $p(y=c|\bx_1)$ and $p(y=c|\bx_2)$.

Given the states, we can define SAS+ transitions and actions according to Definitions~\ref{defn:sas-transition} and~\ref{defn:sas-action}.
For Example 2, $\bx =(x_1, x_2, x_3)$ can be transformed to state $s=(x_1, x_2, x_3)$, $Dom(x_1) =\{0,1\}, Dom(x_2)  = \{0,1\}, Dom(x_3) = \{0,1,2\}$.
For an input $\bx=(male, 2, 500)$, the corresponding state is $s=(0,0,0)$.
The action $a$ changing $x_2$ from 2 to 5 can be represented as $\delta^{x_2}_{0 \rightarrow 1} $.
Thus, the resulting state of applying $a$ is $s\oplus a = (0,1,0)$.

\begin{defn} \label{defn:graph}
\noindent{\bf (Action graph)} Given a SOAP problem $\Pi_{soap}=(H, {\bx}^I, c, O)$, the action graph is a graph $G=(\mathcal{F},E)$ where  $\mathcal{F}$ is the set of transformed states and  an edge $(s_{i-1}, s_{i}) \!\in\!E$ if and only if there is an action $a \in O$ such that $s_{i-1} \oplus a = s_{i}$.
The weight for this edge is $w(s_{i-1}, s_{i}) = \pi(a)$.
\end{defn}



The SOAP problem in Definition~\ref{def:OAP}  is equivalent to finding the shortest path on the state space graph $G=(\mathcal{F},E)$ from a given state $s_I$ to a goal state. A node $s$ is a goal state if $p(y=c|s) \ge z$.
Given the training data $\{X, Y\}$, we use a heuristic search to find a \textbf{preferred goal} state for each $\bx \in X$ that $p(y=c|\bx) < z$. For each of such $\bx$, we find a path in the action graph from $s(\bx)$ to a state $s^*$ such that $p(y=c|s^*) \ge z$ while minimizing the cost of the path.

\begin{algorithm}[tp]
\caption{Heuristic search (Input: $G=(\mathcal{F},E)$, $s_I$) \label{algo:non_opt}}
\begin{algorithmic}[1]
\STATE $N_{es} \leftarrow 0$, $s^* \leftarrow NULL$, $g^* \leftarrow \infty$
\STATE MinHeap.push($s_I$), ClosedList $\leftarrow \{\}$
\WHILE{MinHeap is not empty}
\STATE $s \leftarrow$ MinHeap.pop()
\IF{$p(y=c|s) \ge z$ and $g(s) < g^*$}
    \STATE $N_{es} \leftarrow |ClosedList|$, $s^* \leftarrow s$, $g^* \leftarrow g(s)$
\ENDIF
\STATE \textbf{if} {$|ClosedList|-N_{es}>\Delta$ } \textbf{then return} {$s^*$}
\IF{$\bx \notin$ ClosedList and $p(y=c|s) < z$}
    \STATE ClosedList=ClosedList $\cup\{s\}$
    \FOR{each $(s, s')\in E$}
    \STATE MinHeap.push($s'$)
    \ENDFOR
\ENDIF
\ENDWHILE
\RETURN $s^*$
\end{algorithmic}
\end{algorithm}

Algorithm~\ref{algo:non_opt} shows the heuristic search.
The search uses a standard evaluation function $f(s)=g(s) +  h(s)$.
$g(s)$ is the cost of the path leading up to $s$.
Let the path be $s_0 = s_I$, $s_1$, $\cdots$, $s_m = s$, and $s_{i-1} \oplus a_i = s_i$ for $i=1, \cdots, m$, we have $g(s) =\sum_{i=1}^{m} \pi(a_i)$.
We define the \textbf{heuristic function} as $h(s) = \alpha (z - p(y=c|s))$ if 
$p(y=c|s) < z$, otherwise $h(s) = 0$ .

\nop{
$h(s)$ is a \textbf{heuristic function} defined as 
\begin{eqnarray}
h(s) &=&
\begin{cases}
\alpha (z - p(s)), &  \textrm{if}~p(s) < z\\
0, & \textrm{otherwise}
\end{cases}
\end{eqnarray}
}
For any state $s= s_I \oplus a_1 \oplus a_2 \oplus \cdots \oplus a_m$  satisfying $p(y=c|s) < z$, $f(s)= g(s)+  h(s) = \sum_{i=1}^{m}  \pi(a_i) + \alpha (z - p(y=c|s))$.
Since the goal is to achieve $p(y=c|s) \ge z$, $h(s)$  measures  how far $s$ is from the goal.
$\alpha$ is a controlling parameter. In our experiments, $\alpha$ is set to the mean of all the action costs.

Algorithm~\ref{algo:non_opt} maintains two data structures, a min heap and a closed list, and performs the following main steps:
\begin{enumerate}
\setlength{\parskip}{0pt}
\setlength{\parsep}{0pt}
\item Initialize $N_{es}$, $s^*$, and $g^*$ where $N_{es}$ represent the number of expanded states，$s^*$ is the best goal state ever found, and $g^*$ records the cost of the path leading up to $s^*$. Add the initial state  $s_I$ to the min heap (Lines 1-2).
\item Pop the state $s$ from the heap with the smallest $f(s)$ (Line 4).
\item If $p(y=c|s) \ge z$  and  $g(s) < g^*$, update $g^*$, $N_{es}$, and the best goal state $s^*$ (Lines 5-6).
\item If the termination condition ($|ClosedList|-N_{es}>\Delta$) is met, stop the search and return $s^*$ (Line 8).
\item Add $s$ to the closed list and for each edge $(s, s')\!\in\! E$, add $s'$ to the min heap if $s$ is not in the closed list and not a goal state (Lines 10-12). 
\item Repeat from Step 2.
\end{enumerate}

The closed list is implemented as a set with highly efficient hashing-based duplicate detection.
The search terminates when the search has not found a better plan for a long time ($|ClosedList|-N_{es}>\Delta$). We set a large value ($\Delta=10^7$) in our experiments.
Note that Algorithm~\ref{algo:non_opt} does not have to search all states since it will stop the search once a state \bs~ satisfies the termination condition (Line 8).

By the end of the offline phase, for each $\bx \in X$ and the corresponding state $s(\bx)$, we find a preferred goal state $s^*(\bx)$.
For an input $\bx=(male, 2, 500)$ in Example 2, the corresponding initial state is $s=(0,0,0)$.
An optimal solution is $P=(a_1,a_2,a_3)$ where $a_1=\delta^{x_2}_{0 \rightarrow 1}$, $a_2=\delta^{x_3}_{0 \rightarrow 1}$, $a_3=\delta^{x_3}_{1 \rightarrow 2}$, and the preferred goal state is $s=(0,1,2)$.


\subsection{Online SAS+ planning}
\label{sec:online}

Once the offline phase is done, the results can be used to repeatedly solve SOAP instances. We now describe how to handle a new instance $\bx^I$ and find the actionable plan.

In online SAS+ planning,  we will find a number of closest states of $s(\bx^I)$ and use the combination of their goals to construct the goal $s^*(\bx^I)$. This is inspired by the idea of similarity-based learning methods such as k-nearest-neighbor (kNN).
We first define the similarity between two states.

\begin{defn}
\noindent{\bf (Feature similarity)} Given two states $s(x_1, \cdots, x_M)$ and $s'(x'_1, \cdots, x'_M)$, the similarity of the i-th feature variable is defined as:
\begin{itemize}
\item if the i-th feature is categorical, $\xi_i(s, s') = 1$ if $~x_i = ~x'_i$,  otherwise $\xi_i(s, s') = 0$.
\nop{
\begin{eqnarray}
\xi_i(s, s') &=&
\begin{cases}
0 &  if~x_i \neq ~x'_i\\
1 & otherwise
\end{cases}
\end{eqnarray}
}

\item if the i-th feature is numerical, $\xi_i(s, s') = 1 - \frac{|p_i-p'_i|}{n_i-1},$
where $p_i$ and $p'_i$ are the partition index of features $x_i$ and $x'_i$, and
$n_i$ is the number of partitions of the i-th feature.
\end{itemize}
\end{defn}

Note that $\xi_i(s, s') \in [0,1]$.  $\xi_i(s, s') = 1$ means they are in the same partition, while $\xi_i(s, s') = 0$ means they are totally different.

\begin{defn}
\noindent{\bf (State similarity)} The similarity between two  states $s(x_1, \cdots, x_M)$ and $s'(x'_1, \cdots, x'_M)$ is 0 if there exists $i \in [1, M]$, $x_i$ is a hard attribute and $x_i$ and $x'_i$ are not in the same partition. Otherwise, the similarity is
\begin{eqnarray}
sim(s, s') &=&  \frac{\sum_{i=1}^{M} \phi_i  \xi_i(s, s')}{\sum_{i=1}^{M} {\phi_i}},
\end{eqnarray}
where $\phi_i$ is the feature weight in the random forest.
\end{defn}

\nop{
Initially, we set $\theta=1$. $\phi_i$ is the feature weight computed by the random forest model.
Then, for each i-th feature,
\begin{itemize}
\item if it is a hard variable and $x_i$, $x'_i$ are not in the same partition, then  $\theta=0$.
\item if $x_i$ or $x'_i$ is a missing number, then $\lambda_i=0$, otherwise $\lambda_i=1$.
\end{itemize}
\end{defn}
}

Note that $sim(s, s') \in [0,1]$.
A larger $sim(s, s')$ means higher similarity.
Given two vectors $\bx=(male, 2, 500)$ and $\bx'=(male, 6, 800)$ in Example 2, the corresponding states are $\bs=(0, 0, 0)$ and $\bs'=(0, 1, 0)$. Their feature similarities are $\xi_0(s, s') = 1$, $\xi_1(s, s') = 0$, and $\xi_2(s, s') = 1$.
Suppose $\phi_i=1/3$, then $sim(s, s')=2/3$.

Given two vectors $\bx=(male, 2, 500)$ and $\bx'=(female, 2, 500)$, the corresponding states are $\bs=(0, 0, 0)$ and $\bs'=(1, 0, 0)$. 
Since $x_1$ is a hard attribute and $x_1$, $x'_1$ are not in the same parition, $sim(s, s')=0$.

\noindent\textbf{SAS+ formulation.}
Given a SOAP problem $\Pi_{soap}=(H, {\bx}^I, c, O)$, we define a SAS+ problem $\Pi_{sas}=(\mathcal{X}, \mathcal{O}, s_I, S_G)$ as follows:
\begin{itemize}
\item $\mathcal{X}=\{x_1,\cdots, x_M\}$ is a set of state variables. Each variable $x_i$ has a finite domain $Dom(x_i)=n_i$ where $n_i$ is the number of partitions of the $i$-th feature of ${\bx}$.
\item $\mathcal{O}$ is a set of SAS+ actions directly mapped from $O$ in $\Pi_{soap}$.
\item $s_I$ is transformed from ${\bx}^I$ according to Definition~\ref{defn:state-transformation}.
\item Let $(s_1, \cdots, s_K)$ be the $K$ nearest neighbors of $s_I$ ranked by $sim(s, s_j)$, and their corresponding preferred goal states be $(s^*_1, \cdots, s^*_K)$, the goal in SAS+ is $S_G= \{s^*_1, \cdots, s^*_K\}$. $K>0$ is a user-defined integer.
\end{itemize}

In example 2, if we preprocessed three initial states $s_1=(0,0,0)$, $s_2=(0,1,0)$, $s_3=(0,1,1)$, then three preferred goal states $s^*_1=(0,1,2)$, $s^*_2=(0,1,2)$, and $s^*_3=(0,1,2)$ will be found in the offline phase.
In the online phase, given a new input ${\bx}^I=(male, 2, 1200)$, the corresponding state is $s_I=(0,0,1)$.
Suppose $\phi_i=1/3$, then $sim(s_I, s_1)=5/6$, $sim(s_I, s_2)=1/2$, and $sim(s_I, s_3)=2/3$.
If $K=2$, the 2 nearest neighbors of $s_I$ are $s_1$ and $s_3$, and the goal of the SAS+ problem is $S_G= \{s^*_1, s^*_3\}$.



In the online phase, for a given $\bx^I$, we solve a SAS+ instance defined above. In addition to classical SAS+ planning, we also want to minimize the total action costs. Since some existing classical planners do not perform well in optimizing the plan quality, we employ a SAT-based method.

Our method follows the bounded SAT solving strategy, originally proposed in SATPlan~\cite{Kautz92} and Graphplan~\cite{Blum97}.
It starts from a lower bound of makespan (L=1), encodes the SAS+ problem as a weighted partial Max-SAT (WPMax-SAT) instance~\cite{Lu13tist}, and either proves it unsatisfiable or finds a plan while trying to minimize total action costs at the same time.

\nop{
\begin{algorithm}[tp]
\caption{A WPMax-SAT based SAS+ planning (WSP) framework (Input: A SAS+ problem $\Pi_{sas}$ \label{algo:wsp})}
\begin{algorithmic}[1]
\STATE $L \leftarrow 0$
\WHILE{TRUE}
    \STATE $L \leftarrow L+1$
    \STATE encode SAS+ problem as a WPMax-SAT instance with makespan $L$
    \STATE optimally solve the WPMax-SAT instance
    \IF {a plan is found }
        \STATE decode the plan and return
    \ENDIF
\ENDWHILE
\end{algorithmic}
\end{algorithm}
}

For a SAS+ problem $\Pi_{sas}=(\mathcal{X}, \mathcal{O}, s_I, S_G)$, given a makespan $L$, we define a WPMax-SAT problem $\Psi$ with the following variable set $U$ and clause set $C$. The variable set includes three types of variables:
\begin{itemize}
\item Transition variables: $U_{\delta, t}$, $\forall \delta \in \mT$ and $t\in[1,L]$. 
\item Action variables: $U_{a,t}$, $\forall a\in \mO$ and $t \in [1,L]$.
\item Goal variables: $U_{s^*}$, $\forall s^* \in S_G$.
\end{itemize}

Each variable in $U$ represents the assignment of a transition or an action at time $t$, or a goal condition $s^*$.

The clause set $C$ has two types of clauses: soft clauses and hard clauses.
The soft clause set $C^s$ is constructed as:
$C^s = \{ \neg U_{a,t} | \forall a \in \mO~ \textrm{and}~t \in [1,L] \}$.
For each clause $c=\neg U_{a,t} \in C^s$, its weight is defined as $w(c) = \pi(a)$.
For each clause in the hard clause set $C^h$, its weight is $\sum_{c \in C^s} w(c)$ so that it must be true. $C^h$ has the following hard clauses:
\begin{itemize}
\item Initial state: $\forall x, s_I(x)=f$,
$\bigvee_{\forall \delta^x_{f \rightarrow g} \in \mT(x)}{U_{\delta^x_{f \rightarrow g}, 1}}.$

\item Goal state: $\bigvee_{\forall s* \in S_G}{U_{s^*}}$.
It means at leat one goal condition $s^*$ must be true.

\item Goal condition: $\forall s^* \in S_G$, $\forall x, s^*(x)=g$,
$U_{s^*} \rightarrow \bigvee_{\forall \delta^x_{f \rightarrow g} \in \mT(x)}{U_{\delta^x_{f \rightarrow g},L}}$.
If $U_{s^*}$ is true, then for each assignment $s^*(x)=g$, at least one transition changing variable $x$ to value $g$ must be true at time $L$.

\item Progression: $\forall \delta^x_{f \rightarrow g} \in \mT(x)$ and $t\in[1,L-1]$,
$U_{\delta^x_{f \rightarrow g},t} \rightarrow \bigvee_{\forall \delta^x_{g \rightarrow h} \in \mT(x)} {U_{\delta^x_{g \rightarrow h},t+1}}$.

\item Regression: $\forall \delta^x_{f \rightarrow g} \in \mT(x)$ and $t\in[2,L]$,
$U_{\delta^x_{f \rightarrow g},t} \rightarrow \bigvee_{\forall \delta^x_{h \rightarrow f} \in \mT(x)} {U_{\delta^x_{h \rightarrow f},t+1}}$.

\item Mutually exclusive transitions: for each mutually exclusive transitions pair $(\delta_1, \delta_2)$, $t\in [1,L]$,
$\overline{U_{\delta_1,t}} \bigvee \overline{U_{\delta_2, t}}$.

\item Mutually exclusive actions: for each mutually exclusive actions pair $(a_1, a_2)$, $t\in [1,L]$,
$\overline{U_{a_1,t}} \bigvee \overline{U_{a_2, t}}$.

\item Composition of actions: $\forall a\in \mO$ and $t\in[1,L-1]$,
$U_{a,t} \rightarrow \bigwedge_{\forall \delta \in M(a)} {U_{\delta,t}}$.

\item Action existence: for each non-prevailing transition $\delta \in \mT$,
$U_{\delta,t} \rightarrow \bigvee_{\forall a, \delta \in M(a)} {U_{a,t}}$.

\end{itemize}

There are three main differences between our approach and a related work, SASE encoding~\cite{huang:AAAI10, huang2012sas}.
First, our encoding transforms the SAS+ problem to a WPMax-SAT problem aiming at finding a plan with minimal total action costs while SASE transforms it to a SAT problem which only tries to find a satisfiable plan. 
Second, besides transition and action variables, our encoding has extra goal variables since the goal definition of our SAS+ problem is a combination of several goal states while in SASE it is a partial assignment of some variables.
Third, the goal clauses of our encoding contain two kinds of clauses while SASE has only one since the goal definition of ours is more complicated than SASE.

We can solve the above encoding using any of the MaxSAT solvers, which are extensively studied.
Using soft clauses to optimize the plan in our WPMax-SAT encoding is similar to Balyo's work~\cite{balyo2014different} which uses a MAXSAT based approach for plan optimization (removing redundant actions).



\nop{
Our sub-optimal actionable plans can be viewed as individualized feature selection for each input, while existing feature selection methods are population-based. 
For example, one patient may receive an alert because his blood pressure is too high and needs to be lowered. For another  patient, his heart rate and temperature are key factors triggering the alert and need intervention.  The SOAP solution identifies those few features that can most efficiently change the prediction output of a particular instance. 
}

\section{Experimental Results}
\label{sec:exp}


To test the proposed approach (denoted as ``Planning"), in the offline preprocess, $\Delta$ in Algorithm~\ref{algo:non_opt} is set to $10^7$.
In the online search, we set neighborhood size $K=3$ and use WPM-2014-in~\footnote{http://www.maxsat.udl.cat/} to solve the encoded WPMax-SAT instances.
For comparison, we also implement three solvers: 1) An iterative greedy algorithm, denoted as ``Greedy'' which chooses one action in each iteration that increases $p(y=c|s)$ while minimizes the total action costs.
It keeps iterating until there is no more variables to change.
2) A sub-optimal state space method denoted as ``NS''~\cite{qiang2016}.
3) An integer linear programming (ILP) method~\cite{cui2015}, one of the state-of-the-art algorithms for solving the SOAP problem. ILP gives exact optimal solutions.

\begin{table}
\begin{footnotesize}
\begin{center}
\caption{Datasets information and offline preprocess results.}
\label{tb:offline}
\setlength{\tabcolsep}{2pt}
\centering
\begin{tabular}{lcccccc}
\hline
Dataset & N & D & C & T (s) & \#S & $\sum T$ (days) \\
\hline
Credit & 17714 &  14 &  2 &  1.22 &  3.40E+09 &  4.81E+01 \\ 
A1a & 32561 &  123 &  2 &  365.25 &  1.68E+07 &  7.09E+01 \\ 
Australian & 690 &  14 &  2 &  0.06 &  1.14E+08 &  7.34E-02 \\ 
Breast & 683 &  10 &  2 &  2.43 &  7.07E+07 &  1.99E+00 \\ 
Dna scale & 2000 &  180 &  3 &  161.89 &  3.36E+07 &  6.29E+01 \\ 
Heart & 270 &  13 &  2 &  0.35 &  2.07E+08 &  8.37E-01 \\ 
Ionosphere scale & 351 &  34 &  2 &  64.06 &  8.39E+06 &  6.22E+00 \\ 
Liver disorders & 345 &  6 &  2 &  0.05 &  2.33E+05 &  1.40E-04 \\ 
Mushrooms & 8124 &  112 &  2 &  0.01 &  2.05E+03 &  1.80E-07 \\ 
Vowel & 990 &  10 &  11 &  0.15 &  5.96E+08 &  1.06E+00 \\ 
\hline
\end{tabular}
\end{center}
\end{footnotesize}
\end{table}

We test these algorithms on a real-world credit card company dataset (``Credit'') and other nine benchmark datasets from the UCI repository\footnote{https://archive.ics.uci.edu/ml/datasets.html} and the LibSVM website\footnote{http://www.csie.ntu.edu.tw/${\sim}$cjlin/libsvmtools/datasets/} used in ILP's original experiments~\cite{cui2015}.
Information of the datasets is listed in Table~\ref{tb:offline}.
N, D, and C are the number of instances, features, and classes, respectively.
A random forest is built on the training set using the Random Trees library in OpenCV 2.4.9.
GNU C++ 4.8.4 and Python 2.7 run-time systems are used.

In the offline preprocess, we generate all possible initial states and use Algorithm~\ref{algo:non_opt} to find a preferred goal state for each initial state.
For each dataset, we generate problems with the same parameter settings as in ILP experiments.
Specifically, we use a weighted Euclidean distance as the action cost function.
For action $a$ which changes state $\bs=(x_1, \cdots, x_M)$ to $\bs'=(x'_1, \cdots, x'_M)$, the cost is
\begin{equation} \label{eq:new_action_cost}
\pi(a)  = \sum_{j=1}^M {\beta_j (x_j - x'_j)^2},
\end{equation}
where $\beta_j$ is the cost weight on variable $j$, randomly generated in $[1,100]$.
Since the offline preprocess works are totally independent, we can parallelly solve them in a large number of workstation nodes.
We run the offline preprocess parallelly on a workstation with 125 computational nodes.
Each node has a 2.50GHz processor with 8 cores and 64GB memory. 
For each instance, the time limit is set to 1800 seconds.
If the preprocess search does not finish in 1800 seconds, we record the best solution found in terms of net profit and the total search time (1800 seconds).

We show the average preprocessing time (T) on each dataset in seconds and the total number of possible initial states (\#S) in Table~\ref{tb:offline}.
$\sum T$ shows how many days it costs to finish all preprocess works by parallelly solving in 1000 cores.
We can see that even though the total number of preprocessed states are very large, the total preprocess time can be extensively reduced to an acceptable range by parallelly solving.

\begin{figure*}[ht]
    \centering
    \subfigure[Total offline preprocess time]{\label{subfig:offline-time}\includegraphics[scale=0.35]{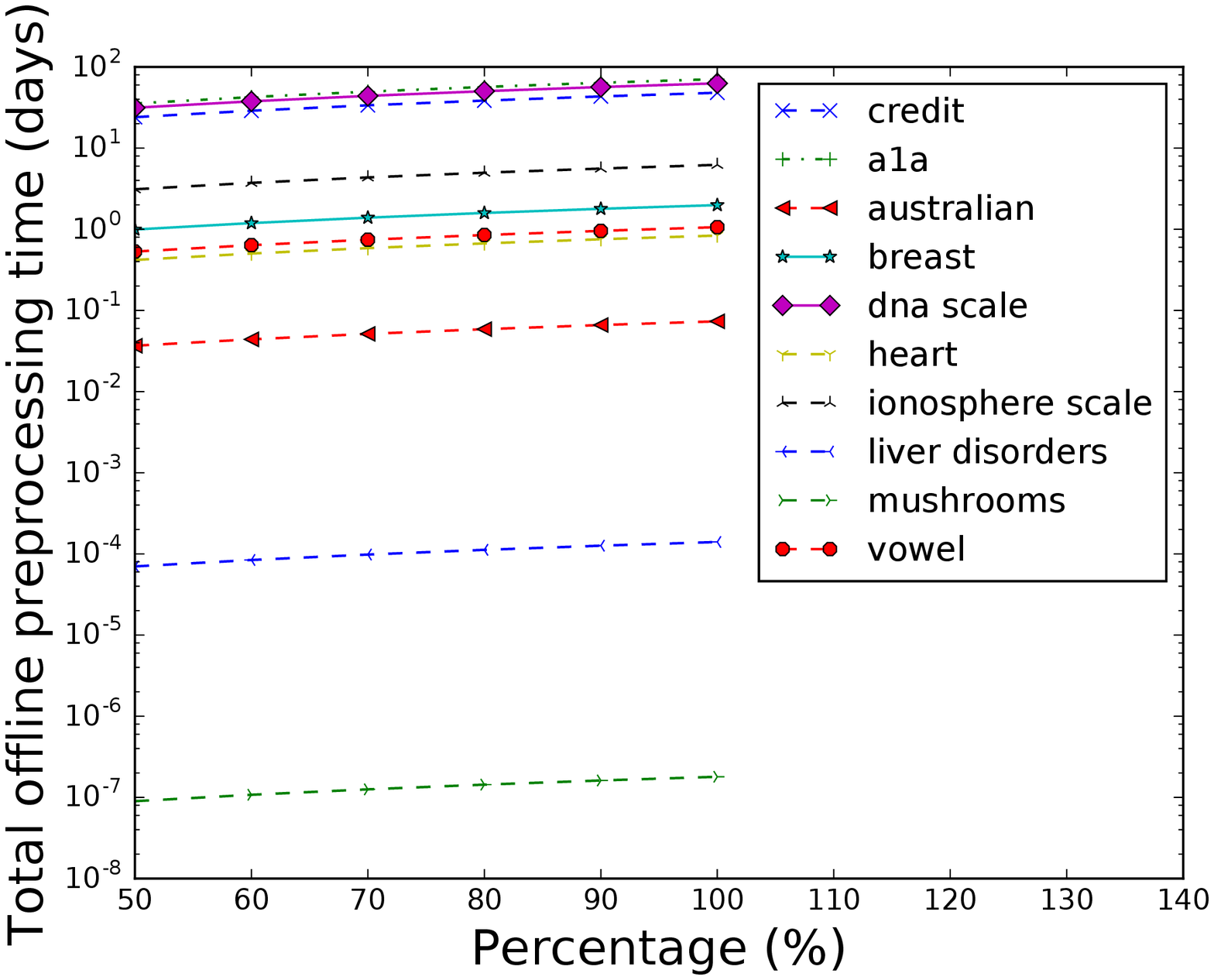}}
    \subfigure[Average online search time]{\label{subfig:offline-online-time}\includegraphics[scale=0.35]{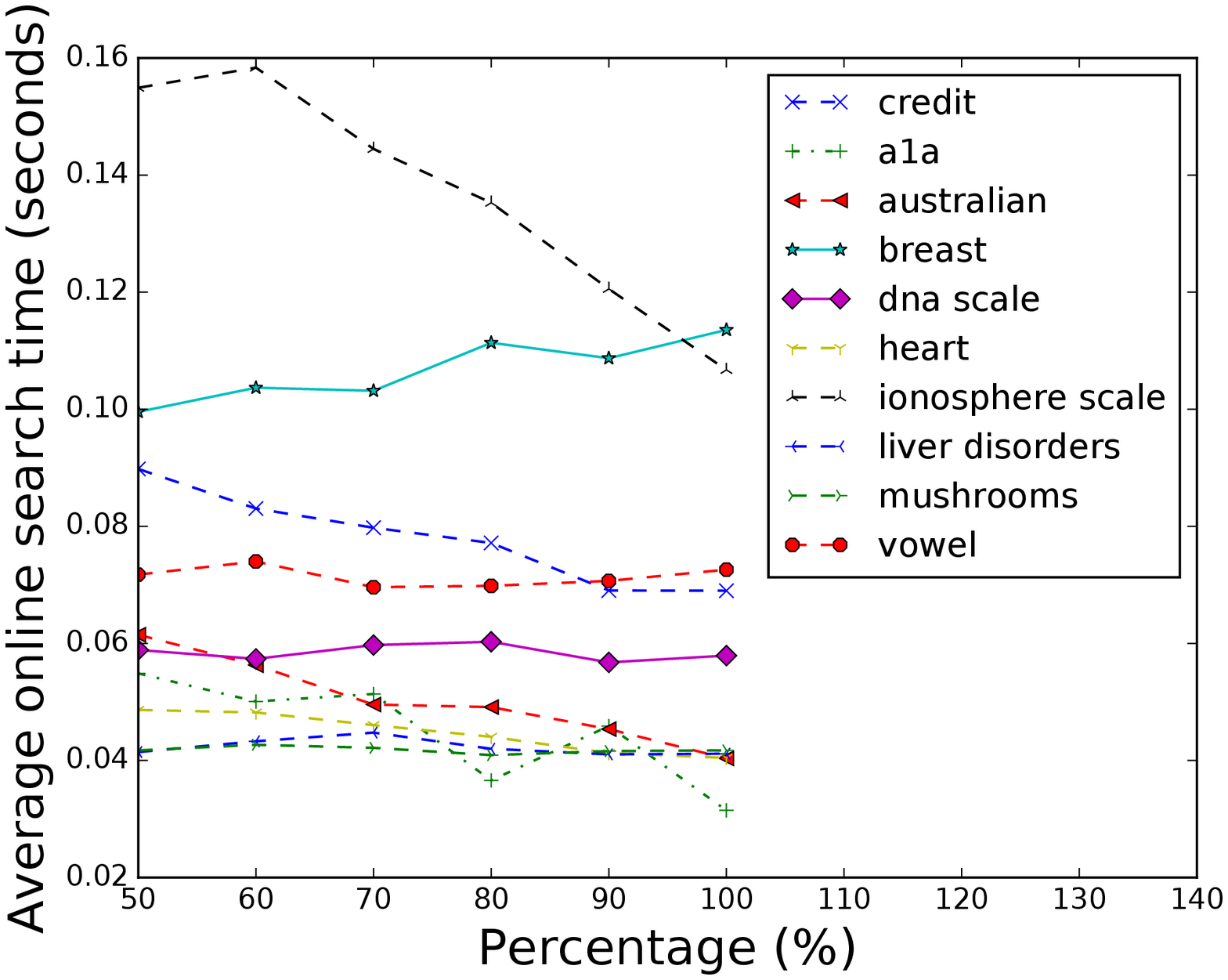}}
    \subfigure[Average total action costs]{\label{subfig:offline-cost}\includegraphics[scale=0.35]{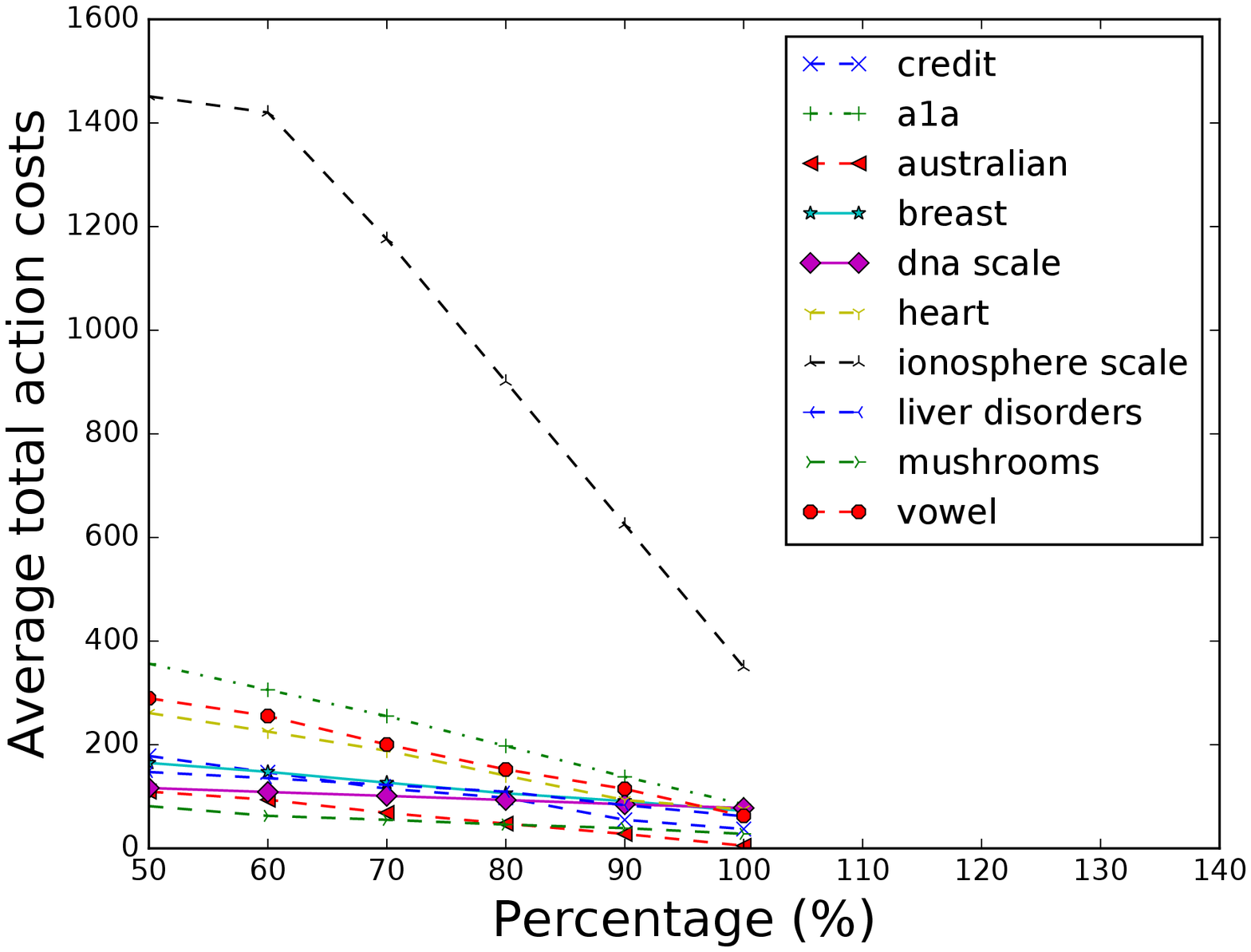}}
    \caption{Experimental results of offline preprocess for different preprocessing percentages}
    \label{fig:offline}
\end{figure*}

In the offline preprocess, the percentage of actual preprocessed states out of all possible initial states in the transformed state space is a key feature of determing the online search quality.
For each preprocessing percentage $r\in (0,100]$ , we randomly sample $r*\#S$ instances from all possible initial states and use Algorithm~\ref{algo:non_opt} to find preferred goals.
Then, in the online search, we randomly sample 100 instances from the test set and generate 100 problems based on these preferred goals.
We report the online search time in seconds and total action costs of the solutions, averaged over 100 runs.
From Figure~\ref{subfig:offline-time}, we can see that the total offline preprocessing time linearly increases with the percentage.
The average total action costs almost linearly decrease with the percentage.
Actually, considering the almost unlimited offline preprocessing time, we can always increase the preprocessing percentage and eventually reach 100\%.

\begin{table*}
\begin{footnotesize}
\begin{center}
\caption{Comparison of four SOAP algorithms on ten datasets. ILP is optimal and others are suboptimal.}
\label{tb:summarize}
\centering
\setlength{\tabcolsep}{1pt}
\begin{tabular}{lccccccccccccccc}
\hline
\multirow{2}{*}{Dataset} & \multicolumn{4}{c}{Greedy}  & \multicolumn{4}{c}{NS} &  \multicolumn{4}{c}{Planning} & \multicolumn{3}{c}{ILP}  \\
\cline{2-16}
&  T (s) & Cost & L & M (GB) & T (s) & Cost & L & M (GB) & T (s) & Cost & L & M (GB) & T (s) & Cost & L \\
\hline
Credit & 1.06 &  525.61 &  12.07 &  0.01 &  1.65 & 33.20 & 3.37 &  0.05 &  \textbf{0.08} & \textbf{33.20} & \textbf{3.17} &  \textbf{15.21} &  6.59 & 33.20 & 3.37 \\ 
A1a & 1.24 &  462.07 &  8.47 &  0.01 &  6.56 & 68.07 & 3.10 &  0.11 &  \textbf{0.05} & \textbf{62.17} & \textbf{3.40} &  \textbf{3.85} &  7.56 & 60.60 & 3.33 \\ 
Australian & 0.04 &  215.10 &  9.30 &  0.01 &  0.06 & 6.03 & 1.37 &  0.01 &  \textbf{0.03} & \textbf{6.03} & \textbf{1.37} &  \textbf{2.98} &  108.89 & 6.03 & 1.37 \\ 
Breast & 0.02 &  375.70 &  16.77 &  0.01 &  0.65 & 74.97 & 11.70 &  0.01 &  \textbf{0.11} & \textbf{74.97} & \textbf{11.70} &  \textbf{1.20} &  30.58 & 74.97 & 11.70 \\ 
Dna scale & 0.11 &  775.26 &  16.68 &  0.01 &  4.59 & 75.30 & 3.00 &  0.08 &  \textbf{0.05} & \textbf{75.30} & \textbf{3.00} &  \textbf{11.26} &  34.54 & 75.30 & 3.00 \\ 
Heart & 0.02 &  569.07 &  9.13 &  0.01 &  0.05 & 83.37 & 2.03 &  0.01 &  \textbf{0.04} & \textbf{83.37} & \textbf{2.03} &  \textbf{5.03} &  5.54 & 83.37 & 2.03 \\ 
Ionosphere scale & 0.04 &  1219.12 &  25.62 &  0.01 & 62.33 & 460.33 & 12.23 &  0.52 &   \textbf{0.13} & \textbf{445.40} & \textbf{12.17} &  \textbf{0.54} &  47.97 & 444.90 & 12.17 \\ 
Liver disorders & 0.04 &  212.67 &  4.90 &  0.01 &  0.07 & 83.17 & 2.50 &  0.01 &  \textbf{0.04} & \textbf{83.17} & \textbf{2.50} &  \textbf{0.01} &  30.47 & 83.17 & 2.50 \\ 
Mushrooms & 0.00 &  58.71 &  1.00 &  0.01 &  0.01 & 30.27 & 1.13 &  0.01 &  \textbf{0.03} & \textbf{30.27} & \textbf{1.13} &  \textbf{0.01} &  3.74 & 30.27 & 1.13 \\ 
Vowel & 0.02 &  425.29 &  9.83 &  0.01 &  0.49 & 61.63 & 4.20 &  0.01 &  \textbf{0.06} & \textbf{61.63} & \textbf{4.20} &  \textbf{11.11} &  66.92 & 61.63 & 4.20 \\ 
\hline
\end{tabular}
\end{center}
\end{footnotesize}
\end{table*}



Table~\ref{tb:summarize} shows a comprehensive comparison in terms of the average search time, the solution quality measured by the total action costs, the action number of solutions, and the memory usage under the preprocessing percentage 100\%.
We report the search time (T) in seconds, total action costs of the solutions (Cost), action number of solutions (L), and the memory usage (GB), averaged over 100 runs.

From Table~\ref{tb:summarize}, we can see that even though our method spends quite a lot of time in the offline processing, its online search is very fast.
Since our method finds near optimal plans for all training samples, its solution quality is much better than Greedy while spending almost the same search time.
Comparing against NP, our method is much faster in online search and maintains better solution qualities in a1a and ionosphere scale and equal solution qualities in other 8 datasets. 
Comparing against ILP, our method is much faster in online search with the cost of losing optimality. 
Typically a trained random forest model will be used for long time. 
Since our offline preprocessing only needs to be run once, its cost is well amortized over large number of repeated uses of the online search. 
In short, our planning approach gives a good quality-efficiency tradeoff: it achieves a near-optimal quality using search time close to greedy search.
Note that since we need to store all preprocessed states and their preferred goal states in the online phase, the memory usage of our method is much larger than greedy and NS approaches.  


\section{Conclusions}
\label{sec:conclusion}

We have studied the problem of extracting actionable knowledge from random forest, one of the most widely used and best off-the-shelf classifiers.
We have formulated the sub-optimal actionable plan (SOAP) problem, which aims to find an action sequence that can change an input instance's prediction label to a desired one with the minimum total action costs.
We have then proposed a SAS+ planning approach to solve the SOAP problem. In an offline phase, we construct an action graph and identify a preferred goal for each input instance in the training dataset. In the online planning phase, for each given input, we formulate the SOAP problem as a SAS+ planning instance based on a nearest neighborhood search on the preferred goals, encode the SAS+ problem to a WPMax-SAT instance, and solve it by calling a WPMax-SAT solver.

Our approach is heuristic and suboptimal, but we have leveraged  SAS+ planning and carefully engineered the system so that it gives good performance. Empirical results on a credit card company dateset and other nine benchmarks have shown that our algorithm achieves a near-optimal solution quality and is ultra-efficient, representing a much better quality-efficiency tradeoff than some other methods.

With the great advancements in data science, an ultimate goal of extracting patterns from data is to facilitate decision making. We envision that machine learning models will be part of larger AI systems that make rational decisions. The support for actionability by these models will be crucial. Our work represents a novel and deep integration of machine learning and planning, two core areas of AI. We believe that such integration will have broad impacts in the future.

Note that the proposed action extraction algorithm can be easily expanded to other additive tree models (ATMs)~\cite{qiang2016}, such as adaboost~\cite{Freund97}, gradient boosting trees~\cite{Friedman01}.
Thus, the proposed action extraction algorithm has very wide applications.

In our SOAP formulation, we only consider actions having deterministic effects. However, in many realistic applications, we may have to tackle some nondeterministic actions.
For instance, push a promotional coupon may only have a certain probability to increase the accumulation effect since people do not always accept the coupon.
We will consider to add nondeterministic actions to our model in the near future.

\acks{This work has been supported in part by National Natural Science Foundation of China (Nos. 61502412, 61033009, and 61175057), Natural Science Foundation of the Jiangsu Province (No. BK20150459), Natural Science Foundation of the Jiangsu Higher Education Institutions (No. 15KJB520036), National Science Foundation, United States (IIS-0534699, IIS-0713109, CNS-1017701), and a Microsoft Research New Faculty Fellowship.}

\end{document}